\RequirePackage{fix-cm}
\documentclass[smallcondensed]{svjour3}     
\usepackage{times}
\usepackage{graphicx}
\usepackage{latexsym}
\usepackage{algorithm}
\usepackage{algpseudocode}
\usepackage{amssymb}
\usepackage{multirow}
\usepackage{url}
\usepackage{array}
\usepackage{float}
\usepackage{longtable}
\usepackage{fullpage} 
\usepackage{comment}
\usepackage[caption = false]{subfig}
\newcolumntype{L}[1]{>{\raggedright\let\newline\\\arraybackslash\hspace{0pt}}m{#1}}
\newcolumntype{C}[1]{>{\centering\let\newline\\\arraybackslash\hspace{0pt}}m{#1}}
\newcolumntype{R}[1]{>{\raggedleft\let\newline\\\arraybackslash\hspace{0pt}}m{#1}}

\begin{document}

\title{Anytime Diagnosis for Reconfiguration}



\author{Alexander Felfernig \and Rouven Walter \and Jos\'{e} A. Galindo \and David Benavides \and Seda Polat Erdeniz  \and M\"usl\"um Atas \and Stefan Reiterer}

\authorrunning{A. Felfernig et al.} 

\institute{Alexander Felfernig \at
             Applied Software Engineering Group, Institute for Software Technology, TU Graz,  Austria \\
              \email{alexander.felfernig@ist.tugraz.at}           
           \and
           Rouven Walter \at
              Symbolic Computation Group, WSI Informatics, Universit\"at T\"ubingen, Germany \\
              \email{rouven.walter@uni-tuebingen.de}           
           \and
           Jos\'{e} A. Galindo \at 
            Computer Languages and Systems Department, University of Sevilla, Spain
              \email{jagalindo@us.es}           
           \and
           David Benavides \at
              Computer Languages and Systems Department, University of Sevilla, Spain \\
              \email{benavides@us.es}           
           \and
           Seda Polat-Erdeniz \at
              Applied Software Engineering Group, Institute for Software Technology, TU Graz,  Austria \\
              \email{spolater@ist.tugraz.at}           
           \and
           M\"usl\"um Atas\at
              Applied Software Engineering Group, Institute for Software Technology, TU Graz,  Austria \\
              \email{muatas@ist.tugraz.at}           
           \and 
Stefan Reiterer \at
               SelectionArts,  Austria \\
              \email{stefan.reiterer@selectionarts.com }           
}

\date{Preprint, cite as: A. Felfernig. R. Walter, J. Galindo, D. Benavides, M. Atas, S. Polat-Erdeniz, and S. Reiterer. Anytime Diagnosis for Reconfiguration. Journal of Intelligent Information Systems, vol. 51, pp. 161-182, 2018.}
\journalname{Journal of Intelligent Information Systems}

\maketitle
\bibliographystyle{ecai2014}

\begin{abstract} Many domains require scalable algorithms that help to determine diagnoses efficiently and often within predefined time limits. \emph{Anytime diagnosis} is able to determine solutions in such a way and thus is especially useful in real-time scenarios such as production scheduling, robot control, and communication networks management where diagnosis and corresponding \emph{reconfiguration} capabilities play a major role. Anytime diagnosis in many cases comes along with a trade-off between diagnosis quality and the efficiency of diagnostic reasoning. In this paper we introduce and analyze \textsc{FlexDiag}  which is an anytime direct diagnosis approach. We evaluate the algorithm with regard to performance and diagnosis quality using a configuration benchmark from the domain of feature models and an industrial configuration knowledge base from the automotive domain. Results show that \textsc{FlexDiag} helps to significantly increase the performance of direct diagnosis search with corresponding quality tradeoffs in terms of minimality and accuracy.
\end{abstract}


\section{Introduction}\label{Introduction}

Knowledge-based configuration is one of the most successful application areas of Artificial Intelligence \cite{felfernighotzbagleytiihonen2014,frayman1987,saphaag,hvam2008,Sabin1998,SalvadorForza2007,Stumptner}. There exist many application domains ranging from \emph{telecommunication systems} \cite{ffhs98b,cocos1994}, \emph{railway interlocking systems} \cite{SiemensStudy}, the \emph{automotive domain} \cite{SinzKaiserKuechlin:03,varisales,WalterKuechlin2014}, \emph{software product lines} \cite{benavides10} to the \emph{configuration of services} \cite{ServiceConfiguration}. 

Configuration technologies must be able to deal with inconsistencies which can occur in different contexts. \emph{First}, a \emph{configuration knowledge base} can be inconsistent, i.e., no solution can be determined. In this context, the task of knowledge engineers is to figure out which constraints are responsible for the unintended behavior of the knowledge base. Bakker et al.  \cite{paperBakker1993} show how to apply model-based diagnosis \cite{Reiter1987} to determine minimal sets of constraints in a knowledge base that are responsible for a given inconsistency. A variant thereof is documented in Felfernig et al. \cite{felfernig2004} where an approach to the automated debugging of  knowledge bases with test cases is introduced. Test cases are interpreted as positive or negative examples that describe the intended behavior of a knowledge base. If some positive examples induce conflicts in the configuration knowledge base, some of the constraints in the knowledge base are faulty and have to be adapted or deleted. If some negative examples are accepted (i.e., not rejected) by the configuration knowledge base, further constraints have to be included in order to take these examples into account (in Felfernig et al. \cite{felfernig2004} this issue is solved by simply including negative examples in negated form into the configuration knowledge base). A related approach in the area of software product lines is proposed in \cite{benavides10-diagnosis}. \emph{Second}, \emph{customer requirements} can be inconsistent with the underlying knowledge base.\footnote{Requirements are additional constraints not part of the configuration knowledge base itself -- these constraints represent specific preferences regarding product properties.} Felfernig et al. \cite{felfernig2004} also show how to diagnose customer requirements that are inconsistent with a configuration knowledge base. The underlying assumption is that the configuration knowledge base itself is consistent but combined with a set of  requirements is inconsistent.

The so far mentioned configuration-related diagnosis approaches are based on conflict-directed hitting set determination where conflicts have to be calculated in order to be able to derive one or more corresponding diagnoses \cite{cr91,Janota2014,junker04quickxplain,Reiter1987,Shah2011}. These approaches often determine diagnoses in a breadth-first search manner which allows the identification of \emph{minimal cardinality diagnoses}. The major disadvantage of applying these approaches is the need of determining minimal conflicts which is inefficient especially in cases where only the \emph{leading diagnoses} (the most relevant ones) are sought. Furthermore, in many application domains it is not necessarily the case that minimal cardinality diagnoses are the preferred ones -- Felfernig et al. \cite{felfernig2009} show how recommendation technologies \cite{JannachZankerFelfernigFriedrich2010} can be exploited for guiding the search for preferred (minimal but not necessarily minimal cardinality) diagnoses.

Algorithms based on the idea of \emph{anytime diagnosis} are useful in scenarios where diagnoses have to be provided in real-time, i.e., within given time limits. Efficient diagnosis and reconfiguration of \emph{communication networks} is crucial to retain the quality of service, i.e., if some components/nodes in a network fail, corresponding substitutes and extensions have to be determined immediately \cite{SimulationReconfiguration,Stumptner99reconfiguration}.  In today's production scenarios which are characterized by small batch sizes and high product variability, it is increasingly important to develop algorithms that support the efficient \emph{reconfiguration of schedules}. Such functionalities support the paradigm of \emph{smart production}, i.e., the flexible and efficient production of highly variant products. Further applications are the diagnosis and repair of \emph{robot control software} \cite{SteinbauerRobocup2005}, sensor networks \cite{pc99},  feature models \cite{Janota2014,benavides10-diagnosis}, the \emph{reconfiguration of cars} \cite{WalterKuechlin2013}, and the \emph{reconfiguration of buildings} \cite{Friedrich2011}. In the diagnosis approach presented in this paper, we assure \emph{diagnosis determination within certain time limits} by systematically reducing the number of solver calls needed. This specific interpretation of \emph{anytime diagnosis} requires a \emph{trade-off} between diagnosis quality (evaluated, e.g., in terms of minimality) and the time needed for diagnosis determination.

Algorithmic approaches to provide efficient solutions for diagnosis problems are manyfold. Some approaches focus on improvements of Reiter's original hitting set directed acyclic graph (HSDAG) \cite{Reiter1987} in terms of a personalized computation of \emph{leading diagnoses} \cite{deKleerAIJournal1990} or other extensions that make the basic approach \cite{Reiter1987} more efficient \cite{Wotawa2001}. Wang et al. \cite{Wang2009} introduce an approach to derive binary decision diagrams (BDDs) \cite{Andersen2010,Bryant1992} on the basis of a pre-determined set of conflicts -- diagnoses can then be determined by finding paths in the BDD that include given variable settings (e.g., requirements defined by the user). A predefined set of conflicts can also be compiled into a corresponding linear optimization problem \cite{Fijany2004}; diagnoses can then be determined by solving the given problem. In knowledge-based recommendation scenarios, diagnoses for user requirements can be pre-compiled in such a way that for a given set of customer requirements, the diagnosis search task can be reduced to querying a relational table (see, for example, \cite{JannachContent2006,Schubert2011}). All of the mentioned approaches either extend the approach of Reiter \cite{Reiter1987} or improve efficiency by exploiting pre-generated information about conflicts or diagnoses. 

An alternative to  conflict-directed diagnosis \cite{Reiter1987} are \emph{direct diagnosis} algorithms that determine minimal diagnoses without the need of predetermining minimal conflict sets \cite{felfernig2012,kostya2014}. The \textsc{FastDiag} algorithm \cite{felfernig2012} is a divide-and-conquer based algorithm that supports the determination of diagnoses without a preceding conflict detection. Such direct diagnosis approaches are especially useful in situations where not the complete set of diagnoses has to be determined but users are interested in the \emph{leading diagnoses}, i.e., diagnoses with a high probability of being relevant for the user. Also in the context of SAT solving,  algorithms have been developed that allow the efficient determination of diagnoses (also denoted as minimal correction subsets) in an efficient fashion \cite{Bacchus2014,Gregoire2014,Silva2013,Mencia2015}. Beside efficiency, prediction quality of a diagnosis algorithm is a major issue in interactive configuration settings, i.e., those diagnoses have to be identified that are relevant for the user. A corresponding comparison of approaches to determine \emph{preferred minimal diagnoses} and unsatisfied clauses with \emph{minimum total weights} is provided in \cite{Walter2016}. The authors point out theoretical commonalities and prove the reducibility of both concepts to each other.

In this paper we show how the \textsc{FastDiag} approach can be converted into an anytime diagnosis algorithm (\textsc{FlexDiag}) that allows tradeoffs between diagnosis quality (minimality and accuracy) and performance. In this paper we focus on \emph{reconfiguration scenarios} \cite{Friedrich2011,SimulationReconfiguration,Stumptner99reconfiguration,WalterKuechlin2014}, i.e., we show how \textsc{FlexDiag} can be applied in situations where a given configuration (solution) has to be adapted conform to a changed set of customer requirements. Our contributions in this paper are the following. First, based on previous work on the diagnosis of inconsistent knowledge bases, we show how to solve reconfiguration tasks with direct diagnosis. Second, we make direct diagnosis anytime-aware by including a parametrization that helps to systematically limit the number of consistency checks and thus make diagnosis search more efficient. Finally, we report the results of a \textsc{FlexDiag}-related evaluation conducted on the basis of real-world configuration knowledge bases (feature models and configuration knowledge bases from the automotive industry) and discuss quality properties of related diagnoses not only in terms of minimality but also in terms of accuracy.

The remainder of this paper is organized as follows. In Section \ref{WorkingExample} we introduce an example configuration knowledge base from the domain of resource allocation. This knowledge base will serve as a working example throughout the paper.  Thereafter (Section \ref{Reconfiguration}) we introduce a definition of a reconfiguration task. In Section \ref{DirectDiagnosis} we discuss basic principles of direct diagnosis on the basis of \textsc{FlexDiag} and show how this algorithm can be applied in reconfiguration scenarios. In Section \ref{Evaluation} we present the results of an analysis of algorithm performance and the quality of determined diagnoses. A simple example of the application of \textsc{FlexDiag} in production environments is given in Section \ref{SmartProduction}. In Section \ref{FutureWork} we discuss issues for future work. With Section \ref{Conclusions} we conclude the paper.

\section{Example Configuration Knowledge Base}\label{WorkingExample}

A configuration system determines configurations (solutions) on the basis of a given set of customer requirements \cite{KnowledgeRepresentation}. In many cases, constraint satisfaction problem (CSP) representations are used for the definition of a configuration task.\footnote{A Constraint Satisfaction Problem (CSP) is typically defined by a set of variables, corresponding finite domains, and a set of constraints \cite{Mackworth1977}. For the representation of constraints we use a notation typically used in the context of CSP solving -- for details see, for example, \cite{felfernighotzbagleytiihonen2014}.} A configuration task and a corresponding configuration (solution) can be defined as follows:

\emph{Definition 1 (Configuration Task and Configuration)}. A configuration task can be defined as a CSP ($V,D,C \cup R$) where $V=\{v_1, v_2, ..., v_n\}$ is a set of variables, $D = \bigcup_{v_i \in V}   \{dom(v_i)\}$ represents domain definitions, and $C = \{c_1, c_2, ..., c_m\}$ is a set of constraints (the configuration knowledge base). Additionally, user requirements are represented by a set of constraints $R = \{r_1, r_2, ..., r_k\}$ where $R$ and $C$ are disjoint. A configuration (solution) for a configuration task is a complete set of assignments (constraints) $S = \{s_1: v_1 = a_1, s_2: v_2 = a_2, ..., s_n: v_n = a_n\}$ where $a_i \in dom(v_i)$ which is consistent with  $C \cup R$.

An example of a  configuration task  represented as a constraint satisfaction problem (CSP) is the following.

\emph{Example (Configuration Task)}.  In this resource allocation problem example, items (a \emph{barrel of fuel}, a \emph{stack of paper}, a \emph{pallet of fireworks}, a \emph{pallet of personal computers}, a \emph{pallet of computer games}, a \emph{barrel of oil}, a \emph{pallet of roof tiles}, and a \emph{pallet of rain pipes}) have to be assigned to three different containers. There are a couple of constraints ($c_i$) to be taken into account, for example, fireworks must not be combined with fuel ($c_1$). Furthermore, there is one requirement ($r_1$) which indicates that the pallet of fireworks has to be assigned to container 1. On the basis of this configuration task definition, a configurator can determine a configuration (solution) $S$.

\begin{itemize}
\item $V = \{fuel, paper, fireworks, pc, games, oil, roof, pipes\}$
\item  \begin{flushleft}$dom(fuel) = dom(paper) = dom(fireworks) = dom(pc) = dom(games) = dom(oil) = dom(roof) = dom(pipes) = \{1,2,3\}$ \end{flushleft}
\item $C = \{c_1: fireworks \neq fuel, c_2: fireworks \neq paper,\\~~~~~~~~~~~~ c_3: fireworks \neq oil,c_4: pipes = roof, c_5: paper \neq fuel\}$
\item $R = \{r_1: fireworks=1\}$
\item $S = \{s_1: pc=3, s_2: games=1, s_3: paper=2, s_4: fuel=3,\\~~~~~~~~~~~~  s_5: fireworks = 1, s_6: oil=2, s_7: roof=1, s_8: pipes =1\}$
\end{itemize}

On the basis of the given definition of a configuration task, we now introduce the concept of \emph{reconfiguration} (see also \cite{Friedrich2011,SimulationReconfiguration,Stumptner99reconfiguration,WalterKuechlin2014}).

\section{Reconfiguration Task}\label{Reconfiguration}

It can be the case that an existing configuration $S$ has to be adapted due to new customer requirements. Examples thereof are changing  requirements  in production schedules, failing components or overloaded network infrastructures in a mobile phone network, and changes in the internal model of the environment of a robot. In the following we assume that the \emph{pallet of paper} should be reassigned to container 3 and the \emph{personal computer} and \emph{games pallets}  should be assigned to the same container.  Formally, the set of \emph{new requirements} is represented by $R_\rho: \{r_1': pc = games, r_2': paper =3\}$. In order to determine reconfigurations, we have to calculate a corresponding diagnosis $\Delta$ (see Definition 2).

\emph{Definition 2 (Diagnosis)}. A diagnosis $\Delta$ (correction subset) is a subset of $S = \{s_1: v_1 = a_1, s_2: v_2 = a_2, ..., s_n: v_n = a_n\}$ such that $S - \Delta \cup C \cup R_\rho$ is consistent. $\Delta$ is minimal if there does not exist a diagnosis $\Delta'$ with $\Delta' \subset \Delta$.

On the basis of the definition of a minimal diagnosis, we can introduce a formal definition of a reconfiguration task.

\emph{Definition 3 (Reconfiguration Task and Reconfiguration)}. A reconfiguration task can be defined as a CSP ($V,D,C,S,R_\rho$) where $V$ is a set of variables, $D$ represents variable domain definitions,  $C$ is a set of constraints, $S$ represents an existing configuration, and $R_\rho=\{r_1', r_2', ..., r_q'\}$ ($R_\rho$ consistent with $C$) represents a set of reconfiguration requirements. Furthermore, let $\Delta$ be a minimal diagnosis for the reconfiguration task. A reconfiguration is a variable assignment $S_\Delta$ = $\{s_1: v_1 = a_1', s_2: v_2 = a_2', ..., s_l: v_l = a_l'\}$ where $s_i \in \Delta$, $a_i' \neq a_i$, and $S - \Delta \cup S_\Delta \cup C \cup R_\rho$ is consistent.

If $R_\rho$ is inconsistent with $C$, the new requirements have to be analyzed and changed before a corresponding reconfiguration task can be triggered \cite{felfernigAIMAG2011,felfernig2009}. An example of a reconfiguration task in the context of our configuration knowledge base is the following.

\emph{Example (Reconfiguration Task)}.  In the resource allocation problem, the original customer requirements $R$ are substituted by the requirements $R_\rho = \{r_1': pc = games, r_2': paper =3\}$. The resulting reconfiguration task instance is the following.

\begin{itemize}
\item $V = \{fuel, paper, fireworks, pc, games, oil, roof, pipes\}$
\item \begin{flushleft}$dom(fuel) = dom(paper) = dom(fireworks) = dom(pc) = dom(games) = dom(oil) = dom(roof) = dom(pipes) = \{1,2,3\}$\end{flushleft}
\item $C = \{c_1: fireworks \neq fuel, c_2: fireworks \neq paper,\\~~~~~~~~~~~~ c_3: fireworks \neq oil, c_4: pipes = roof, c_5: paper \neq fuel\}$
\item $S = \{s_1: pc=3, s_2: games=1, s_3: paper=2, s_4: fuel=3,\\~~~~~~~~~~~~ s_5: fireworks = 1, s_6: oil=2, s_7: roof=1, s_8: pipes =1\}$
\item $R_\rho = \{r_1': pc = games, r_2': paper =3\}$
\end{itemize}

To solve a reconfiguration task (see Definition 3), conflict-directed diagnosis approaches \cite{Reiter1987} would determine a set of minimal conflicts and then determine a hitting set that resolves each of the identified conflicts. In this context, a minimal conflict set $CS \subseteq S$ is a minimal set of variable assignments that trigger an inconsistency with $C \cup R_\rho$, i.e., $CS \cup C \cup R_\rho$ is inconsistent and there does not exist a conflict set $CS'$ with $CS' \subset CS$. In our working example, the minimal conflict sets are $CS_1:\{s_1: pc = 3, s_2: games = 1\}$, $CS_2: \{s_3: paper = 2\}$, and $CS_3: \{s_4: fuel = 3\}$. The corresponding minimal diagnoses are $\Delta_1: \{s_1, s_3, s_4\}$ and $\Delta_2: \{s_2, s_3, s_4\}$. 

The elements in a diagnosis indicate which variable assignments have to be adapted such that a reconfiguration can be determined that takes into account the new requirements in $R_\rho$. Consequently, a reconfiguration represents a minimal set of changes to the original configuration ($S$) such that the new requirements $R_\rho$ are taken into account. If we choose $\Delta_1$, a reconfiguration $S_\Delta$ (reassignments for the variable assignments in $\Delta_1$) can be determined by a CSP solver call $C \cup R_\rho \cup (S-\Delta_1)$. The resulting configuration $S'$ can be $\{s_1: pc=1, s_2: games=1, s_3: paper=3, s_4: fuel=2, s_5: fireworks = 1, s_6: oil=2, s_7: roof=1, s_8: pipes =1\}$. For a detailed discussion of conflict-based diagnosis we refer to Reiter \cite{Reiter1987}. In the following we introduce an approach to the determination of minimal reconfigurations  which is based on a \emph{direct diagnosis} algorithm, i.e., diagnoses are determined without the need of determining related minimal conflict sets.

\section{Reconfiguration with \textsc{FlexDiag}}\label{DirectDiagnosis}

In the following discussions, the set $AC = C \cup R_\rho \cup S$ represents the union of all constraints that restrict the set of possible solutions for a given reconfiguration task. Furthermore, $S$ represents a set of constraints that are considered as candidates for being included in a diagnosis $\Delta$. The idea of \textsc{FlexDiag} (Algorithm 1) is to systematically filter out the constraints that become part of a minimal diagnosis using a divide-and-conquer based approach.

\begin{algorithm}[ht]
\caption{}\label{flexdiag}
\begin{algorithmic}[1]
\Procedure{FlexDiag}{$S, AC=C \cup R_\rho \cup S,m$}{:~$\Delta$}
\If{isEmpty($S$) or inconsistent($AC-S$)} \State{return $\emptyset$;}
\Else \State{return \textsc{FlexD($\emptyset$,$S$,$AC$,$m$)};}
\EndIf
\EndProcedure
\Statex
\Procedure{FlexD}{$D,S=\{s_1..s_q\},AC,m$}{:~$\Delta$}
\If{$D \neq \emptyset$ and consistent ($AC$)} \State{return $\emptyset$;}
\ElsIf{size($S$) $\leq$ $m$} \State{return $S$;}
\EndIf
\State{$k=\lceil \frac{q}{2} \rceil$;}
\State{$S_1 = \{s_1..s_k\}; S_2 = \{s_{k+1}..s_q\}$;}
\State{$D_1 = \textsc{FlexD}(S_1,S_2,AC-S_1,m)$;}
\State{$D_2 = \textsc{FlexD}(D_1,S_1,AC-D_1,m)$;}
\State{$return(D_1 \cup D_2)$;}
\EndProcedure
\end{algorithmic}
\end{algorithm}

\emph{Sketch of Algorithm}. In our example reconfiguration task, the original configuration $S = \{s_1, s_2, s_3,$ $s_4, s_5, s_6, s_7, s_8\}$ and the new set of customer requirements is $R_\rho = \{r_1', r_2'\}$. Since $S \cup R_\rho \cup C$ is inconsistent, we are in need of a minimal diagnosis $\Delta$ and a reconfiguration $S_\Delta$ such that $S - \Delta \cup S_\Delta \cup R_\rho \cup C$ is consistent. In the following we will show how the \textsc{FlexDiag} (Algorithm 1) can be applied to determine such a minimal diagnosis $\Delta$.

\begin{figure*}
\centering
\includegraphics[height=6.5cm]{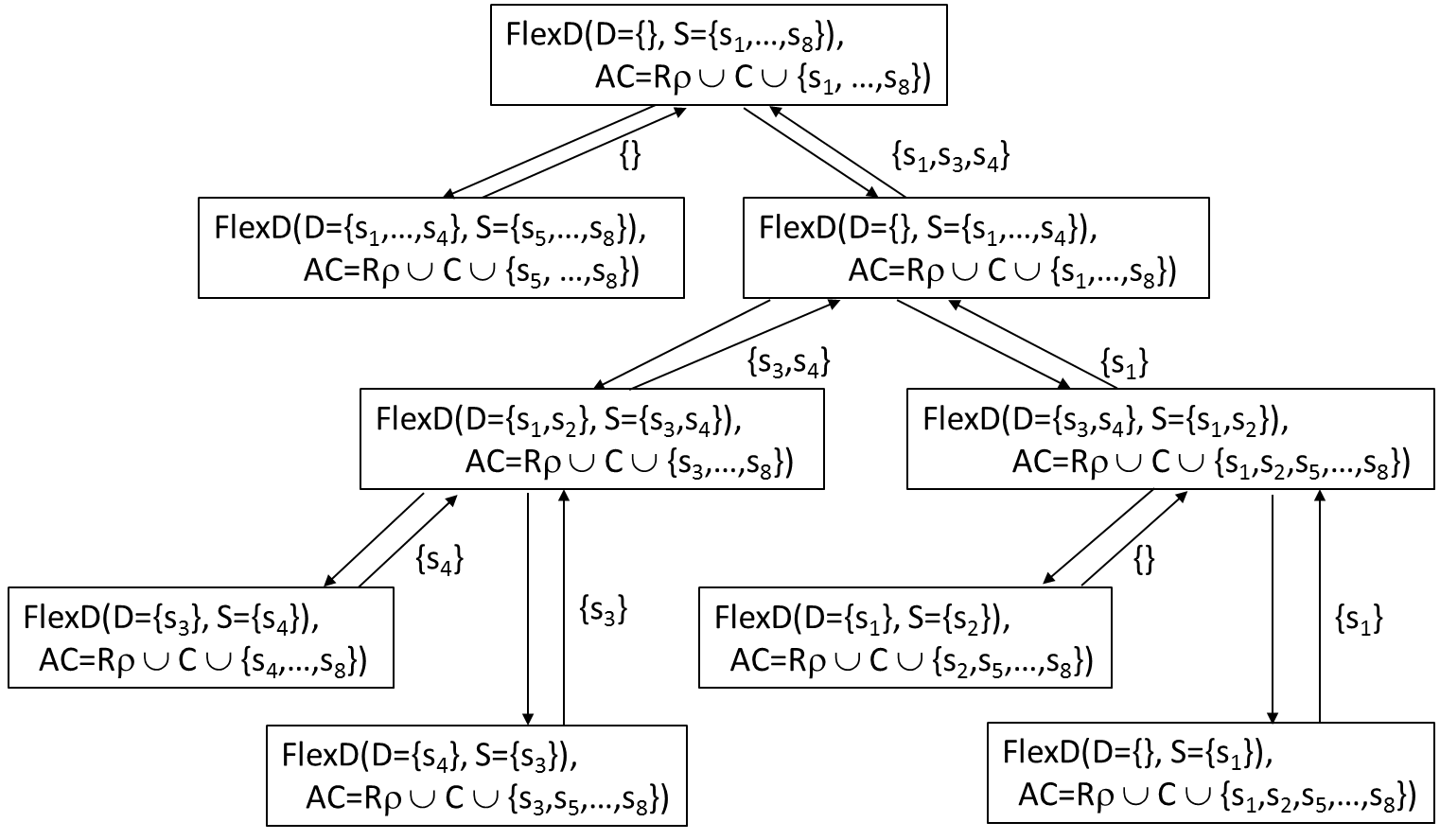}
\caption{\textsc{FlexDiag} walkthrough: determining one minimal diagnosis with $m=1$ ($\Delta = \{s_1, s_3, s_4\}$).}
\label{fig:flexd1}
\end{figure*}

\begin{figure*}
\centering
\includegraphics[height=4.25cm]{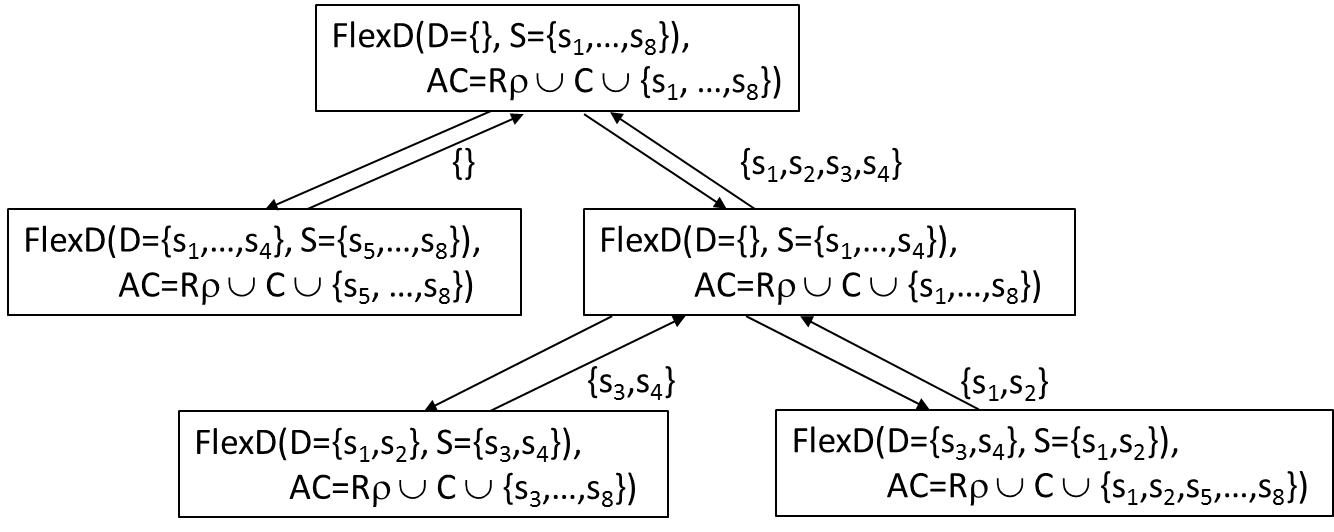}
\caption{\textsc{FlexDiag} walkthrough: determining a minimal diagnosis with $m=2$ ($\Delta = \{s_1, s_2, s_3, s_4\}$).} 
\label{fig:flexd2}
\end{figure*}

\textsc{FlexDiag} is assumed to be activated under the assumption that $AC=C \cup R_\rho \cup S$ is inconsistent, i.e., the consistency of $AC$ is not checked by the algorithm. If $AC$ is inconsistent but $AC - S$ is also inconsistent, \textsc{FlexDiag} will not be able to identify a diagnosis in $S$; therefore $\emptyset$  is returned. Otherwise, a recursive function \textsc{FlexD} is activated which is in charge of determining one minimal diagnosis $\Delta \subseteq S$. In each recursive step, the  constraints in $S$ are divided into two different subsets ($S_1$ and $S_2$) in order to figure out if already one of these subsets includes a diagnosis. If this is the case, the second set must not be inspected for diagnosis elements anymore. If we assume, for example, $S = \{s_1, s_2, s_3, s_4, s_5, s_6, s_7, s_8\}$ is inconsistent and we divide $S$ into the two subsets $S_1 = \{s_1, s_2, s_3, s_4\}$ and $S_2 = \{s_5, s_6, s_7, s_8\}$ and $S_1$ is already consistent with $C \cup R_\rho$ then diagnosis elements are searched in $S_2$ (since $S_1$ is already consistent). The complete related walkthrough is depicted in Figures \ref{fig:flexd1} and \ref{fig:flexd2}.

\textsc{FlexDiag} is based on the concepts of \textsc{FastDiag} \cite{felfernig2012}, i.e., it returns one diagnosis ($\Delta$) at a time and is complete in the sense that if a diagnosis is contained in $S$, then the algorithm will find it. A corresponding reconfiguration can be determined by a CSP solver call $C \cup R_\rho \cup (S-\Delta)$. The determination of multiple diagnoses at a time can be realized on the basis of the construction of a HSDAG \cite{Reiter1987}. In \textsc{FlexDiag}, the parameter $m$ is used to control \emph{diagnosis quality} in terms of \emph{minimality}, \emph{accuracy}, and the \emph{performance} of diagnostic search (see Section \ref{Evaluation}). The higher the value of $m$ the higher the performance of \textsc{FlexDiag} and the lower the degree of diagnosis quality. The inclusion of $m$ to control quality and performance is the major difference between \textsc{FlexDiag} and \textsc{FastDiag}. If  $m = 1$ (see Algorithm 1), the number of consistency checks needed for determining one minimal diagnosis is $2\delta \times log_2 (\frac{n}{\delta})+2\delta$ (in the worst case) \cite{felfernig2012}. In this context, $\delta$ represents the set size of the minimal diagnosis $\Delta$ and $n$ represents the number of constraints in solution $S$.

If $m > 1$, the number of needed consistency checks can be systematically reduced if we accept the tradeoff of possibly loosing the property of diagnosis minimality (see Definition 2).  If we allow settings with $m>1$, we can reduce the upper bound of the number of consistency checks to  $2\delta \times log_2(\frac{2n}{\delta \times m})$ (in the worst case). These upper bounds regarding the number of needed consistency checks allow to estimate the worst case runtime performance of the diagnosis algorithm which is extremely important for \emph{real-time scenarios}. Consequently, if we are able to estimate the upper limit of the time needed for completing one consistency check (e.g., on the basis of simulations with an underlying constraint solver), we are also able to figure out lower bounds for $m$ that must be chosen in order to guarantee a \textsc{FlexDiag} runtime within predefined time limits.  

Table \ref{numberconsistencychecks} depicts an overview of consistency checks needed depending on the setting of the parameter $m$ and the diagnosis size $\delta$ for $|S|=16$. For example, if $m=2$ and the size of a minimal diagnosis is $\delta = 4$, then the upper bound for the number of needed consistency checks is 16. If the size of $\delta$ further increases, the number of corresponding consistency checks does not increase anymore. Figures \ref{fig:flexd1} and \ref{fig:flexd2} depict \textsc{FlexDiag} search trees depending on the setting of granularity parameter $m$. The upper bound for the number of consistency checks helps us to determine the maximum amount of time that will be needed to determine a diagnosis on the basis of \textsc{FlexDiag}. For example, if the maximum time needed for one consistency check is $20 ms$, the maximum time needed for determining a diagnosis with $m=2$ (given $\delta = 8$) is $\approx 320$ milliseconds.

\begin{table}[ht]
\footnotesize
\centering{}\begin{tabular}{|c|c|c|c|c|c|c|c|c|c|c|c|c|c|c|c|} 
\hline
$\delta$ & m=1 & m=2 & m=4 &  m=8  \\
\hline
1 & 10 & 8 & 6 & 4 \\
\hline
2 & 16 & 12 & 8 & 4 \\
\hline
4 & 24 & 16 & 8 & - \\
\hline
8 & 32 & 16 & - & - \\
\hline
16 & 32 & - & - & - \\
\hline
\end{tabular}   \caption{Worst-case estimates for the number of needed consistency checks depending on the granularity parameter $m$ and the diagnosis size $\delta$ for $|S|=16$.}  \label{numberconsistencychecks}
\end{table}

\textsc{FlexDiag}  determines one diagnosis at a time which indicates variable assignments of the original configuration that have to be changed such that a reconfiguration conform to the new requirements ($R_\rho$) is possible. The algorithm supports the determination of \emph{leading diagnoses}, i.e., diagnoses that are preferred with regard to given user preferences \cite{felfernig2012,Walter2016}. \textsc{FlexDiag} is based on a strict lexicographical ordering of the constraints in $S$: the lower the importance of a constraint $s_i \in S$ the lower the index of the constraint in $S$. For example, $s_1: pc = 3$ has the lowest ranking. The lower the ranking, the higher the probability that the constraint will be part of a reconfiguration $S_\Delta$. Since $s_1$ has the lowest priority and it is part of a conflict, it is element of the diagnosis returned by \textsc{FlexDiag}. For a discussion of the properties of lexicographical orderings we refer to \cite{felfernig2012,junker04quickxplain}.

\section{Evaluation}\label{Evaluation}
In this section, we present the evaluation we executed to verify the performance of \textsc{FlexDiag}. We first analyze how \textsc{FlexDiag} performs in front of real and randomly generated models and then, compare it with an evolutionary approach. 

\subsection{Evaluation aspects}
To evaluate \textsc{FlexDiag}, we analyzed the two aspects of (1) \emph{algorithm performance} (in terms of \emph{milliseconds} needed to determine one minimal diagnosis) and (2) \emph{diagnosis quality} (in terms of \emph{minimality} and \emph{accuracy} -- see Formulae \ref{minimality} and \ref{accuracy}). We analyzed both aspects by varying the value of parameter $m$. Our hypothesis in this context was that the higher the value of $m$, the lower the number of needed consistency checks (the higher the efficiency of diagnosis search) and the lower diagnosis quality in terms of the share of diagnosis-relevant constraints returned by \textsc{FlexDiag}.  Diagnosis quality can, for example, be measured in terms of the degree of \emph{minimality} of the constraints in a diagnosis $\Delta$ (see Formula \ref{minimality}), i.e., the cardinality of $\Delta$ compared to the cardinality of $\Delta_{min}$. $|\Delta_{min}|$ represents the cardinality of a minimal diagnosis identified with $m=1$.

\begin{equation}\label{minimality}
minimality(\Delta)=\frac{|\Delta_{min}|}{|\Delta|}
\end{equation}

If $m>1$, there is no guarantee that the diagnosis $\Delta$ determined for $S$ is a superset of the diagnosis $\Delta_{min}$ determined for $S$ in the case $m=1$. Besides minimality, we introduce \emph{accuracy} as an additional quality indicator (see Formula \ref{accuracy}). The higher the share of elements of $\Delta_{min}$ in $\Delta$, the higher the corresponding accuracy (the algorithm is able to reproduce the elements of the minimal diagnosis for $m=1$).

\begin{equation}\label{accuracy}
accuracy(\Delta)=\frac{|\Delta \cap \Delta_{min}|}{|\Delta_{min}|}
\end{equation} 

\subsection{Datasets and Results}
We evaluated \textsc{FlexDiag} with regard to both metrics (algorithm performance, and diagnosis quality) by applying the algorithm to different benchmarks. First, using random feature models generated with the Betty tool \cite{betty}. Second, with the set of models hosted in the S.P.L.O.T repository\footnote{\url{www.splot-research.org}}. Third, in a real-world model extracted from the last Ubuntu Linux distribution \cite{galindo10-acota}. Finally, a real-world automotive dataset. The configuration models are feature models which include requirement constraints, compatibility constraints, and different types of structural constraints such as mandatory relationships and alternatives.

For all the different datasets we report on averaged values. For that, we first, calculate the acurracy, execution time, and minimality for all the executions. Then, we aggregate the data and calculate the mean for the metrics.  

\subsubsection{Experimental platform} The experiments were conducted using a version of \textsc{FlexDiag} implemented in Java and integrated in the FaMa Tool Suite \cite{benavides2013fama}. All the models were translated to a Constraint Satisfaction Problem (CSP) and used the Choco library for consistency checking.\footnote{\url{www.choco-solver.org}}  Further, our \textsc{FlexDiag} implementation was running in a grid of computers running on four-CPU Dell Blades with Intel Xeon X5560 CPUs running at 2.8GHz, with 8 threads per CPU, and CentOS v6. The total RAM memory was 8GB. To parallelize the executions we used GNU Parallel \cite{Tange2011a}.

\subsubsection{Random models}
The first dataset used to evaluate \textsc{FlexDiag} was randomly generated. We used BeTTy~\cite{betty} to generate a dataset that ranges from 50 to 2000 features and 10\% to 30\% of cross-tree constraints. The generation approach is based on Th{\"u}m \emph{et al.} \cite{thum2009reasoning} that imitates realistic topologies.

For each model combining a given number of features and a percentage of cross-tree constraints, we randomly generated different sizes of reconfiguration requirements that involved the 10\%, 30\%, 50\% and 100\% of features of the model. Then we randomly reordered each of the reconfiguration requirements 10 times (to prevent ordering biases). Moreover, we executed \textsc{FlexDiag} on each combination of parameters three times to get average execution times trying to avoid third party threads. 

In the following, we present the results showing a comparison between the different values of \textit{m} and how the values evolved depending on the size of the models. Note that to generate the plots we aggregated the data and therefore the values shown are averaged results.

\begin{figure}[htpb]
\centering
\includegraphics[width = 1\textwidth]{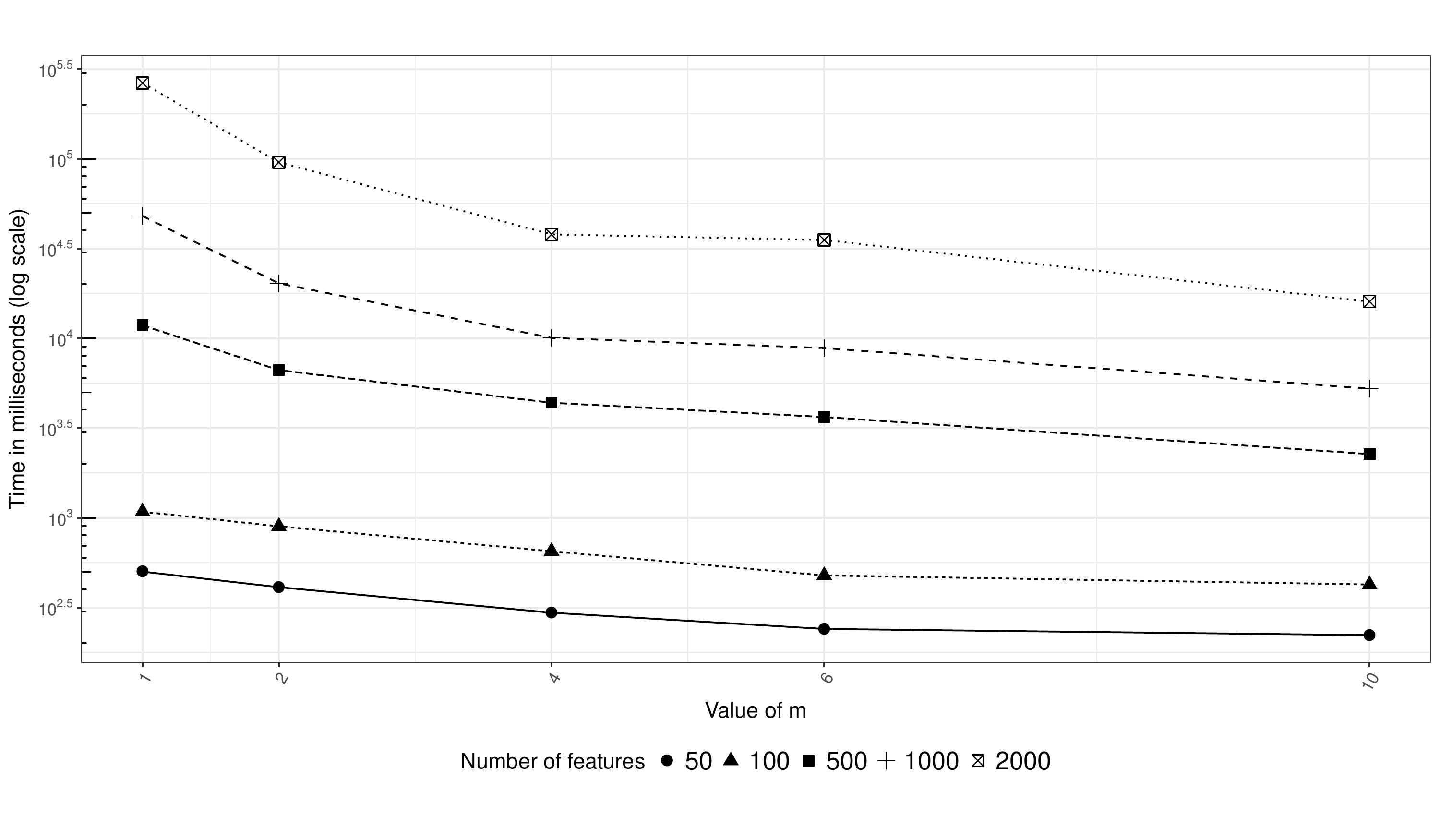}
\includegraphics[width = 1\textwidth]{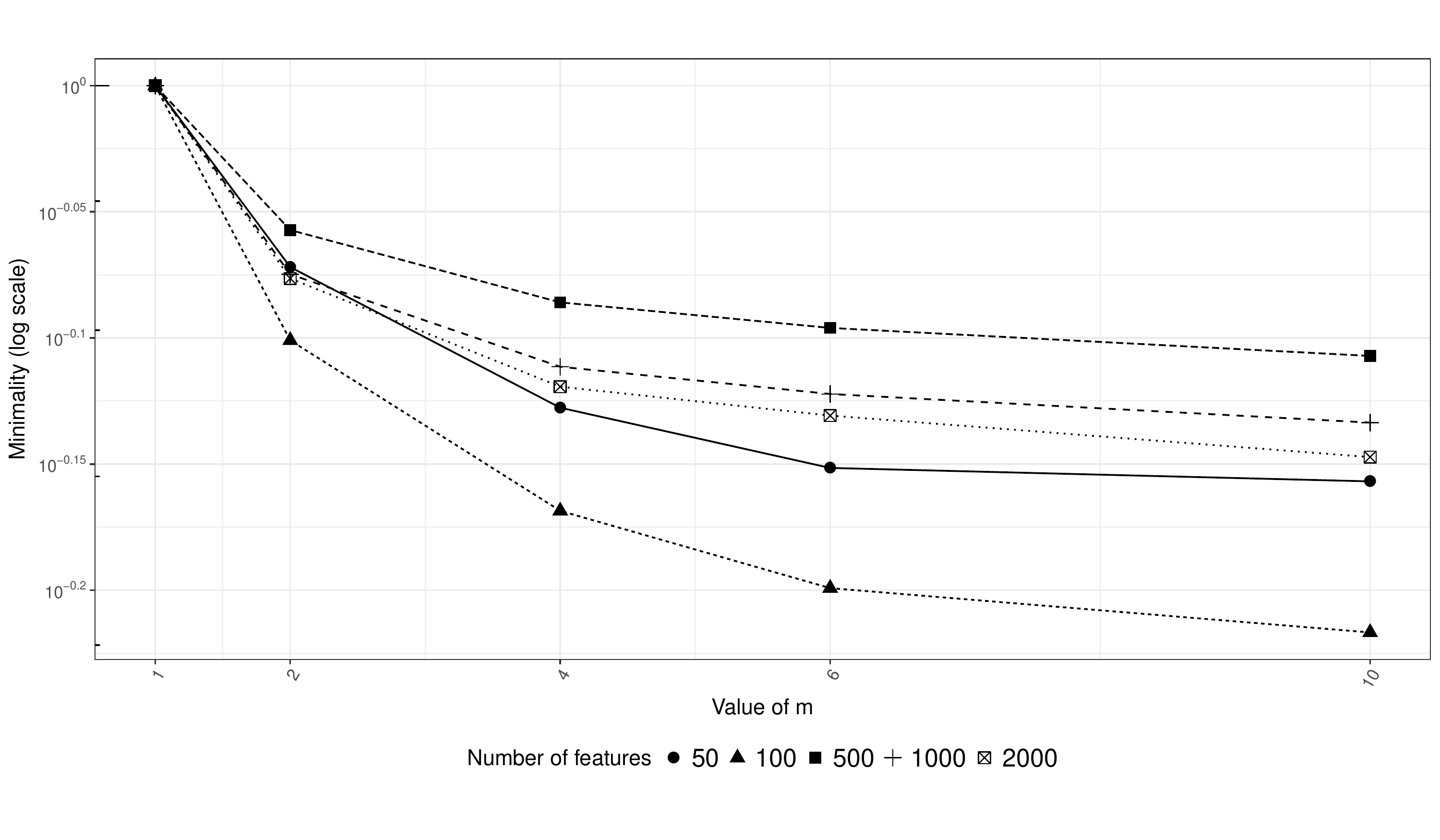}
\caption{Random evolution based on features, time and \textit{m}}
\label{fig:random}
\end{figure}

Figure \ref{fig:random} shows how the diagnosis performance can be increased depending on the setting of the \textit{m} parameter. Also we observe how the minimality deteriorates when increasing \textit{m}.

\begin{table}[htpb]
\centering
\caption{Random evaluation depending on \textit{m} value and model size.  $|V|$  represents the number of variables in the CSP. The second column shows the $m$ value used in \textsc{FlexDiag}. $|C|$ refers to the number of constraints in the model, $|\Delta|$ is the average size of a diagnosis, and \emph{average time} (in milliseconds), \emph{average minimality}, and \emph{average accuracy} represent the means of the calculated values for those metrics.\label{tab:random}}
\begin{tabular}{ccccccc}
  \hline
$|V|$ & m & $|C|$ & $|\Delta|$  & Average time & Average minimality & Average accuracy \\ \hline \hline 

  \hline
 \multirow{5}{*}{50} &   1 &  \multirow{5}{*}{58.40} & 14.62 & 501.63 & 1.00 & 1.00 \\ 
  &   2 &  & 15.72 & 410.92 & 0.85 & 0.92 \\ 
  &   4 &  & 17.22 & 296.14 & 0.75 & 0.87 \\ 
  &   6 &  & 18.71 & 240.52 & 0.71 & 0.87 \\ 
  &  10 &  & 19.07 & 222.15 & 0.70 & 0.86 \\ \hline
  \multirow{5}{*}{100} &   1 &  \multirow{5}{*}{109.80} & 27.51 & 1081.24 & 1.00 & 0.99 \\ 
  &   2 &  & 32.80 & 896.99 & 0.79 & 0.90 \\ 
  &   4 &  & 38.19 & 651.20 & 0.68 & 0.88 \\ 
  &   6 &  & 40.99 & 477.64 & 0.63 & 0.87 \\ 
  &  10 &  & 42.29 & 425.16 & 0.61 & 0.88 \\ \hline
  \multirow{5}{*}{500} &   1 &  \multirow{5}{*}{566.40} & 182.62 & 11808.41 & 1.00 & 1.00 \\ 
  &   2 &  & 203.77 & 6647.27 & 0.88 & 0.96 \\ 
  &   4 &  & 219.41 & 4372.98 & 0.82 & 0.95 \\ 
  &   6 &  & 221.82 & 3643.81 & 0.80 & 0.96 \\ 
  &  10 & & 231.49 & 2265.53 & 0.78 & 0.97 \\ \hline
  \multirow{5}{*}{1000} &   1 &  \multirow{5}{*}{1141.00} & 347.25 & 47956.27 & 1.00 & 1.00 \\ 
    &   2 &  & 398.62 & 20227.16 & 0.84 & 0.97 \\ 
    &   4 &  & 434.76 & 10051.52 & 0.77 & 0.96 \\ 
    &   6 &  & 440.64 & 8825.81 & 0.75 & 0.96 \\ 
    &  10 &  & 461.53 & 5242.34 & 0.74 & 0.98 \\ \hline
   \multirow{5}{*}{2000} &   1 &  \multirow{5}{*}{2274.80} & 663.36 & 264084.35 & 1.00 & 1.00 \\ 
    &   2 &  & 743.21 & 95467.42 & 0.84 & 0.97 \\ 
    &   4 &  & 818.12 & 37911.86 & 0.76 & 0.97 \\ 
    &   6 &  & 828.39 & 35271.86 & 0.74 & 0.96 \\ 
    &  10 &  & 888.38 & 16000.95 & 0.71 & 0.97 \\ 
   \hline
\end{tabular}
\end{table}

Table \ref{tab:random} shows the averaged data we obtained. It is worth mentioning that the minimality decreases when \textit{m} increases and that accuracy still provides acceptable results with \textit{m} = 10. Also, the execution time (in milliseconds) is less than five minutes in the worst case. 

As we can observe in Table \ref{tab:random} and Figure \ref{fig:random}, while the execution time decreases when incrementing $m$, quality deteriorates. However, minimality is clearly affected while accuracy stays with minor variations. Also, we observe that the time improvement depends on $m$ and the number of features. For example, if we compare the time between $m=1$ and $m=10$, we can increase in runtime of 2.26$\times$ for 50 features, 5.21$\times$ for 500 features, 9.14$\times$ for 1000 and 16.5$\times$ for 2000 features.

\subsubsection{SPLOT repository models}
We extracted a total of 387 models from the SPLOT repository. For each model, we randomly generated different sizes of reconfiguration requirements that involved the 10\%, 30\%, 50\% and 100\% of features of the model. Then we randomly reordered each of the reconfiguration requirements 10 times (to prevent ordering biases). Moreover, we executed \textsc{FlexDiag} on each combination of parameters three times to get average execution times trying to avoid third party threads. 

Table \ref{tab:splotModels} shows the data of those models categorized as realistic in the repository. We again see that \textsc{FlexDiag} scales with no problem offering a good trade-off between accuracy and minimality while keeping the average runtime (in milliseconds) below a second. 


\begin{table}[htpb]
\centering
\caption{\textsc{FlexDiag} results in front of SPLOT realistic models.  $|V|$  represents the number of variables in the CSP. The second column shows the $m$ value used in \textsc{FlexDiag}. \emph{Model Name} is the model name in SPLOT. $|C|$ refers to the number of constraints in the model, $|\Delta|$ is the average size of a diagnosis, and \emph{average time} (in milliseconds), \emph{average minimality}, and \emph{average accuracy} represent the means of the calculated values for those metrics. \label{tab:splotModels}
}
\begin{tabular}{ccp{1.7cm}ccccc}
  \hline
  $|V|$ & m & Model Name &$|C|$ & $|\Delta|$ & Average time & Average minimality & Average accuracy \\ \hline \hline
    \multirow{5}{*}{290} &   1 &\multirow{5}{*}{ REAL-FM-4} & \multirow{5}{*}{61.00} & 1.00 & 527.45 & 1.00 & 1.00 \\ 
    &   2 &  &  & 1.75 & 527.83 & 0.62 & 1.00 \\ 
    &   4 &  &  & 2.75 & 464.33 & 0.48 & 1.00 \\ 
   &   6 & &  & 4.00 & 444.65 & 0.40 & 1.00 \\ 
   &  10 & &  & 6.25 & 432.65 & 0.35 & 1.00 \\ \hline
\multirow{5}{*}{88} &   1 & \multirow{5}{*}{REAL-FM-1} & \multirow{5}{*}{26.00} & 6.93 & 566.92 & 1.00 & 1.00 \\ 
 &   2 &  &  & 10.48 & 514.35 & 0.62 & 0.86 \\ 
 &   4 &  &  & 15.02 & 411.55 & 0.42 & 0.76 \\ 
 &   6 &  &  & 21.17 & 316.52 & 0.32 & 0.70 \\ 
 &  10 &  &  & 22.58 & 317.92 & 0.28 & 0.70 \\ \hline
\multirow{5}{*}{44} &   1 & \multirow{5}{*}{REAL-FM-20} & \multirow{5}{*}{8.00} & 3.50 & 225.32 & 1.00 & 0.99 \\ 
 &   2 &  &  & 5.35 & 201.23 & 0.60 & 0.86 \\ 
 &   4 &  &  & 8.07 & 171.22 & 0.38 & 0.84 \\ 
 &   6 &  &  & 10.15 & 155.45 & 0.31 & 0.80 \\ 
 &  10 &  &  & 10.90 & 152.78 & 0.28 & 0.80 \\ \hline
\multirow{5}{*}{23} &   1 & \multirow{5}{*}{REAL-FM-2} & \multirow{5}{*}{9.00} & 2.78 & 177.55 & 1.00 & 1.00 \\ 
 &   2 &  &  & 4.07 & 161.72 & 0.68 & 0.86 \\ 
 &   4 &  &  & 4.93 & 144.05 & 0.54 & 0.82 \\ 
 &   6 &  &  & 6.60 & 131.78 & 0.41 & 0.76 \\ 
 &  10 &  &  & 7.62 & 216.98 & 0.36 & 0.76 \\ \hline
\multirow{5}{*}{43} &   1 & \multirow{5}{*}{REAL-FM-3} & \multirow{5}{*}{13.00} & 1.75 & 211.97 & 1.00 & 1.00 \\ 
 &   2 &  &  & 3.03 & 199.62 & 0.60 & 0.95 \\ 
 &   4 &  &  & 4.78 & 178.35 & 0.36 & 0.93 \\ 
 &   6 &  &  & 7.38 & 161.82 & 0.24 & 0.92 \\ 
 &  10 &  &  & 9.15 & 147.82 & 0.21 & 0.88 \\ \hline
\end{tabular}
\end{table}

As we can observe in Table \ref{tab:splotModels}, while the execution time decreases when incrementing \emph{m}, quality deteriorates again. However minimality is clearly affected while accuracy stays with minor variations, although, we can observe some special cases ("REAL-FM-5" with $m=2$) were it deteriorates a bit more.

\subsubsection{Ubuntu-based model}

In order to test \textsc{FlexDiag} with large-scale real models, we encoded the variability existing in the \emph{Debian} packaging system for the \emph{Ubuntu} distribution and generated a set of configurations representing Ubuntu user installations with wrong package selections. Concretely, we modelled the Ubuntu Xenial \footnote{\url{http://releases.ubuntu.com/16.04/}} distribution containing 58,107 packages and 52,721 constraints. This model was extracted using the mapping presented in \cite{galindo10-acota,galindo10-fmsple}. We executed \textsc{FlexDiag} with different \textit{m} values. We randomly generated different sizes of reconfiguration requirements that involved the 10\%, 30\%, 50\% and 100\% of features of the model. Then we randomly reordered each of the reconfiguration requirements 10 times (to prevent ordering biases). Moreover, we executed \textsc{FlexDiag} on each combination of parameters three times to get average execution times trying to avoid third party threads.


Table \ref{tab:xenial} shows that \textsc{FlexDiag} is able to provide a good accuracy even with \textit{m} set to 10. Also, it shows as expected, the negative impact of \textit{m} regarding minimality. We observe that the execution time with $m=1$ was $3.7$ hours while with $m=10$ it was $2.5$ hours. This represents an improvement of runtime in 1.47$\times$.

\begin{table}[htpb]
\centering
\caption{Results obtained after executing \textsc{FlexDiag} with the Ubuntu Xenial variability model. The seconds column shows the $m$ value used in \textsc{FlexDiag}. $|V|$  represents the number of variables in the CSP.  $|C|$ refers to the number of constraints in the CSP, $|\Delta|$ is the average size of a diagnosis, and \emph{average time} (in milliseconds), \emph{average minimality}, and \emph{average accuracy} represent the means of the calculated values for those metrics.\label{tab:xenial}}
\begin{tabular}{rrccccc}
  \hline 
  $|V|$ & m  &$|C|$ & $|\Delta|$  & Average time & Average minimality & Average accuracy \\ \hline \hline
 \multirow{5}{*}{ 58107} &   1  & \multirow{5}{*}{105459} & 1.75 & 13523986.27 & 1.00 & 1.00 \\ 
   &   2 & & 3.40 & 12245231.77 & 0.51 & 0.75 \\ 
   &   4 &  & 6.12 & 10800288.33 & 0.31 & 0.71 \\ 
   &   6 &  & 7.15 & 10546545.60 & 0.24 & 0.71 \\ 
   &  10 &  & 12.48 & 9208147.25 & 0.14 & 0.71 \\ 
\hline
\end{tabular}
\end{table}

\subsubsection{Automotive models}
The benchmark used in this experiment includes three automotive configuration models from a German car manufacturer. 
For each model, we randomly generated different sizes of reconfiguration requirements that involved the 10\%, 30\%, 50\% and 100\% of features of the model. Then we randomly reordered each of the reconfiguration requirements 10 times (to prevent ordering biases). Moreover, we executed \textsc{FlexDiag} on each combination of parameters three times to get average execution times trying to avoid third party threads. 

\begin{table}[ht]
\centering
\caption{\textsc{FlexDiag} evaluated with benchmarks from the automotive industry  (calculation of the first diagnosis: $ids$ 1--3 represent different type series of a German premium car manufacturer). The second column shows the $m$ value used in \textsc{FlexDiag}. $|V|$  represents the number of variables in the CSP.  $|C|$ refers to the number of constraints in the CSP, $|\Delta|$ is the average size of a diagnosis, and \emph{average time} (in milliseconds), \emph{average minimality}, and \emph{average accuracy} represent the means of the calculated values for those metrics. \label{tab:evaluations3}}
\begin{tabular}{lrrrrccc}
  \hline
id & m & $|V|$ & $|C|$ & $|\Delta|$ &Average time & Average minimality & Average accuracy \\ 
  \hline \hline
\multirow{5}{*}{01} &   1 & \multirow{5}{*}{1888} & \multirow{5}{*}{7404} & 623.78 & 22812981.47 & 1.00 & 1.00 \\ 
   &   2 &  &  & 787.43 & 6720901.40 & 0.80 & 0.99 \\ 
   &   4 &  &  & 870.25 & 1785352.90 & 0.71 & 0.99 \\ 
   &   6 &  &  & 876.97 & 1716644.85 & 0.70 & 0.99 \\ 
   &  10 &  &  & 892.85 & 473017.90 & 0.69 & 1.00 \\ \hline
 \multirow{5}{*}{02} &   1 &\multirow{5}{*}{ 1828} &\multirow{5}{*}{ 5451} & 611.38 & 8730834.77 & 1.00 & 1.00 \\ 
   &   2 &  &  & 760.85 & 3176414.95 & 0.81 & 0.99 \\ 
   &   4 &  &  & 845.47 & 914621.30 & 0.72 & 0.99 \\ 
   &   6 &  &  & 850.28 & 878283.40 & 0.71 & 1.00 \\ 
   &  10 &  &  & 865.22 & 237203.47 & 0.70 & 1.00 \\ \hline
  \multirow{5}{*}{03} &   1 &\multirow{5}{*}{ 1843} & \multirow{5}{*}{8056} & 369.00 & 22147386.47 & 1.00 & 1.00 \\ 
   &   2 &  &  & 538.50 & 11901657.38 & 0.80 & 0.98 \\ 
   &   4 &  &  & 605.00 & 3381818.66 & 0.70 & 0.99 \\ 
   &   6 &  &  & 613.31 & 3104191.16 & 0.68 & 0.99 \\ 
   &  10 &  &  & 626.59 & 823533.78 & 0.67 & 0.99 \\  \hline
\end{tabular}
\end{table}

Table \ref{tab:evaluations3} shows that \textsc{FlexDiag} is able to provide a good accuracy even with \textit{m} set to 10. Again, it shows as expected, the negative impact of \textit{m} regarding minimality. Also, the execution time for model with $id=1$ and $m=1$ was $6.33$ hours while with $m=10$ it was $7.9$ minutes. This represents an improvement of 26.9$\times$.

\subsection{Comparing \textsc{FlexDiag} with Evolutionary Algorithms}
In this research we do not compare \textsc{FlexDiag} with more traditional diagnosis approaches -- for related evaluations we refer the reader to \cite{felfernig2012} were detailed analyses can be found. These analyses clearly indicate that direct diagnosis approaches outperform standard diagnosis approaches based on the resolution of minimal conflicts \cite{Reiter1987} (if the search goal is to identify not all minimal but the so-called \emph{leading diagnoses} which should, for example,  be shown to users in interactive settings).  

However, in this Section we compare \textsc{FlexDiag}  with an evolutionary algorithm inspired by \cite{cendic2014genetic,li2002computing}. The evolutionary algorithm has been build with the jenetics framework for Java\footnote{\url{http://jenetics.io/}} leaving all parameters default and fixing the process to 500 generations. Also we compare the performance of \textsc{FlexDiag} with \textit{m} set to 1 to have a fair comparison. Note that with higher values of \emph{m}, we would even perform better in terms of runtime.

The first observation is that the evolutionary approach was not capable of dealing with very large realistic models (Ubuntu, automotive) when setting a time-out of 24 hours. Therefore, we report this comparison only relying on randomly generated models. 

\begin{figure}[h!]
\centering
\includegraphics[width = \textwidth]{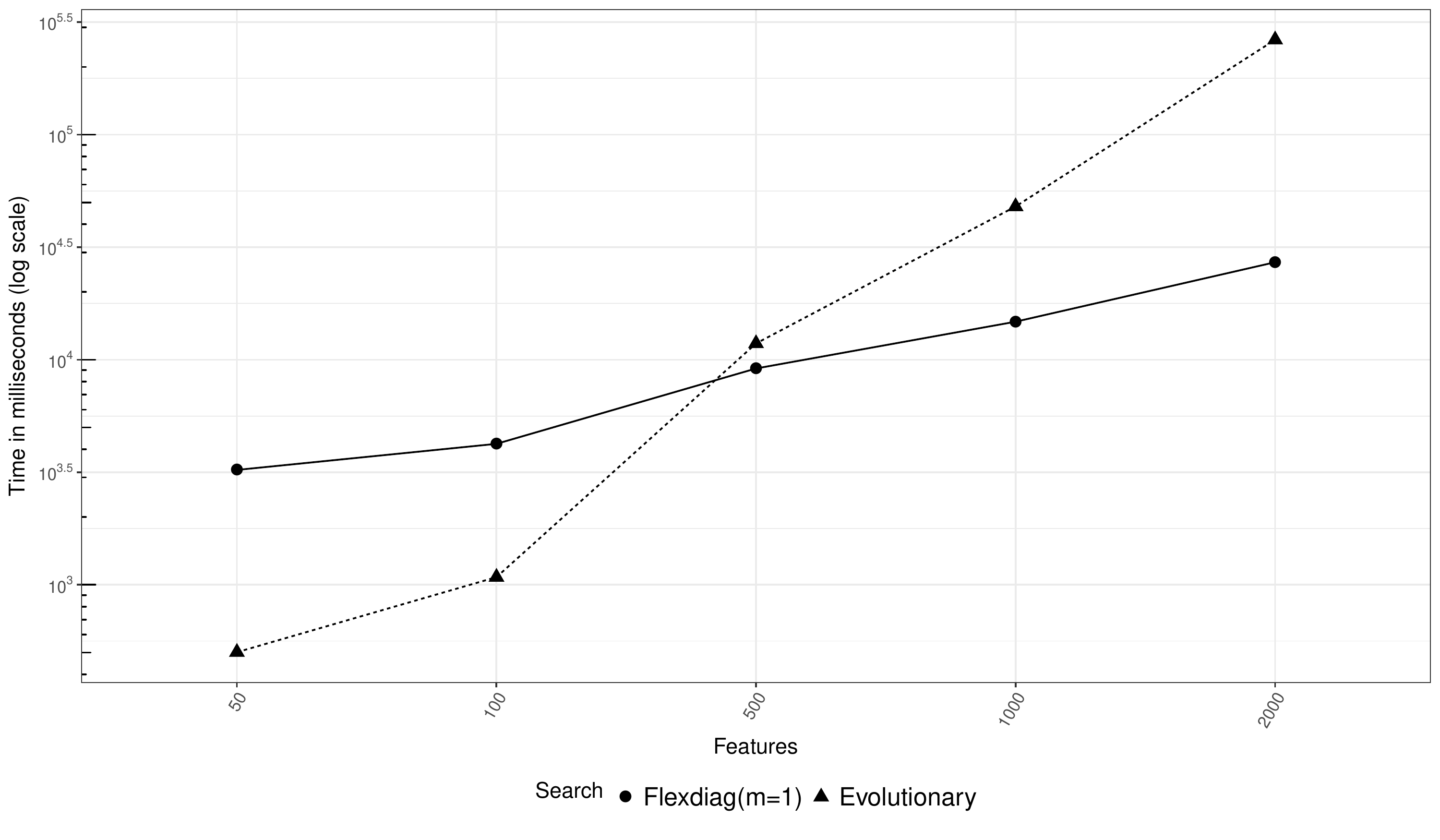}
\caption{Comparison between \textsc{FlexDiag} and the evolutionary approach regarding time}
\label{fig:vs_time}
\end{figure}

Figure \ref{fig:vs_time} shows that the required time for \textsc{FlexDiag} is usually higher for models having less than 500 features. Therefore, there is a point when \textsc{FlexDiag} pays off and scales much better. Also, it is worth mentioning that the evolutionary algorithm was not capable of obtaining a complete and minimal explanation and returned only partial diagnoses. This is, in 500 generations it only  found partial explanations. 

Table \ref{tab:vs} shows that \textsc{FlexDiag} returns minimal diagnoses while we observe that the evolutionary approach was not capable of detecting minimal diagnoses. Also, we do see that \textsc{FlexDiag} offered a much better accuracy. 

\begin{table}[ht]
\centering
\caption{Comparison results of the evolutionary approach and \textsc{FlexDiag}.  $|V|$  represents the number of variables in the CSP. \emph{Approach} refers to the used approach, $|C|$ refers to the number of constraints in the CSP, $|\Delta|$ is the average size of a diagnosis, and \emph{average time} (in milliseconds), \emph{average minimality}, and \emph{average accuracy} represent the means of the calculated values for those metrics.\label{tab:vs}}
\begin{tabular}{clccp{1.6cm}p{1.4cm}p{1.2cm}}
  \hline
    $|V|$ & Approach  &$|C|$  & $|\Delta|$ & Average time & Average minimality & Average accuracy \\ \hline \hline
 50 & Evolutionary & 58.40 & 1.53 & 3237.87 & 8.84 & 0.33 \\ 
 50 & \textsc{FlexDiag} & 58.40 & 14.62 & 501.63 & 1.00 & 1.00 \\ \hline
 100 & Evolutionary & 109.80 & 2.13 & 4229.25 & 10.81 & 0.17 \\ 
 100 & \textsc{FlexDiag} & 109.80 & 27.51 & 1081.24 & 1.00 & 0.99 \\ \hline
 500 & Evolutionary & 566.40 & 9.45 & 9142.43 & 15.82 & 0.03 \\ 
 500 & \textsc{FlexDiag} & 566.40 & 182.62 & 11808.41 & 1.00 & 1.00 \\ \hline
 1000 & Evolutionary & 1141.00 & 22.07 & 14754.50 & 11.49 & 0.03 \\ 
 1000 & \textsc{FlexDiag} & 1141.00 & 347.25 & 47956.27 & 1.00 & 1.00 \\ \hline
 2000 & Evolutionary & 2274.80 & 48.61 & 27111.93 & 9.32 & 0.03 \\ 
 2000 & \textsc{FlexDiag} & 2274.80 & 663.36 & 264084.35 & 1.00 & 1.00 \\ \hline
\end{tabular}
\end{table}

\subsection{Threats to validity}

Even though the experiments presented in this paper provide evidence that the solution proposed is valid, there are some assumptions that we made that may affect their validity. In this section, we discuss the different threats to validity that affect the evaluation.

\textbf{External validity.} The inputs used for the experiments presented in this paper were either realistic or designed to mimic realistic feature models. The Debian feature model and the Automotive are realistic since numerous experts were involved in the design. However, since they  were developed using a manual design process, it may have errors and not encode all configurations. Also, the random feature models may not accurately reflect the structure of real feature models used in industry.   
The major threats to the external validity are: 
\begin{itemize}
  \item \emph{Population validity}, the real feature models that we used may not represent all valid  configurations in the domains due it manual construction. Also, random models might not have the same structure as real models (e.g. mathematical operators used in the complex constraints). To reduce these threats to validity, we generated the models using previously published techniques \cite{thum2009reasoning} and using existing implementations of these techniques in Betty \cite{betty}. 
  \item \emph{Ecological validity}: While external validity, in general, is focused on the generalization of the results to other contexts (e.g. using other models), the ecological validity ii focused on possible errors in the experiment materials and tools used. To prevent ecological validity threats, such as third party threads running in the virtual machines and impacting performance, the \textsc{FlexDiag} analyses were executed three times and then averaged. 
\end{itemize}

\textbf{Internal validity} The CPU resources required to analyse a feature model depend on the number of features and percentage of cross-tree constraints. However, there may be other variables that affect performance, such as the nature of the constraints used. To minimize these other possible effects, we introduced a variety of models to ensure that we covered a large part of the constraint space.

\subsection{Final Remarks}
We observed that \textsc{FlexDiag} scales up with random and real-world feature models. Observing that, generally, diagnosis quality in terms of minimality and accuracy deteriorates with an increasing size of parameter $m$. 

Minimality and accuracy depend on the configuration domain and are not necessarily monotonous. For example, since a diagnosis determined by \textsc{FlexDiag} is not necessarily a superset of a diagnosis determined with $m = 1$, it can be the case that the \emph{minimality} of a diagnosis determined with $m > 1$ is greater than 1 (if \textsc{FlexDiag} determines a diagnosis with lower cardinality than the minimal diagnosis determined with $m=1$). For simplicity, let us assume that $AC = S = \{c_1,c_2,c_3,c_4,c_5,c_6,c_7,c_8\}$ and the following conflict sets $CS_i$ exist between the constraints $c_i \in S$: $CS_1: \{c_1,c_3\}$, $CS_2: \{c_2,c_3\}$, and $CS_3: \{c_4,c_6\}$. Given $m=1$, \textsc{FlexDiag} would determine the diagnosis $\{c_1,c_2,c_4\}$ whereas in the case of $m=2$, $\{c_3,c_4\}$ is returned by the algorithm.

\section{Another Example: Reconfiguration in Production}\label{SmartProduction}

The following simplified reconfiguration task is related to \emph{scheduling in production} where it is often the case that, for example, schedules and corresponding production equipment has to be reconfigured. In this example setting, we do not take into account configurable production equipment (configurable machines) and limit the reconfiguration to the assignment of orders to corresponding machines. The assignment of an order $o_i$ to a certain machine $m_j$ is represented by the corresponding variable $o_im_j$. The domain of each such variable represents the different possible slots in which an order can be processed, for example, $o_1m_1 = 1$ denotes the fact that the processing of order $o_1$ on machine $m_1$ is performed during and finished after time slot 1. 

Further constraints restrict the way in which orders are allowed to be assigned to machines, for example, $o_1m_1 < o_1m_2$ denotes the fact that order $o_1$ must be completed on machine $m_1$ before a further processing is started on machine $m_2$. Furthermore, no two orders must be assigned to the same machine during the same time slot, for example, $o_1m_1 \neq o_2m_1$ denotes the fact that order $o_1$ and $o_2$ must not be processed on the same machine in the same time slot (slots 1..3). Finally, the definition of our reconfiguration task is completed with an already determined schedule $S$ and a corresponding reconfiguration request represented by the reconfiguration requirement $R_\rho = \{r_1': o_3m_3 < 5\}$, i.e., order $o_3$ should be completed within less than 5 time units.

\begin{itemize} 
\item $V = \{o_1m_1, o_1m_2, o_1m_3, o_2m_1, o_2m_2, o_2m_3, o_3m_1, o_3m_2, o_3m_3\}$
\item \begin{flushleft}$dom(o_1m_1) = dom( o_2m_1) = dom(o_3m_1)=\{1,2,3\}.$ $dom(o_1m_2) = dom(o_2m_2) = dom(o_3m_2) = \{2,3,4\}.$  $dom(o_1m_3)  = dom(o_2m_3) = dom( o_3m_3)=\{3,4,5\}.$ \end{flushleft}
\item $C = \{c_1: o_1m_1 < o_1m_2, c_2: o_1m_2 < o_1m_3,\\~~~~~~~~~~~$ $c_3: o_2m_1 < o_2m_2, c_4: o_2m_2 < o_2m_3, $ $c_5: o_3m_1 < o_3m_2,\\~~~~~~~~~~~$ $c_6: o_3m_2 < o_3m_3,$ $ c_7: o_1m_1 \neq o_2m_1, \\~~~~~~~~~~~~c_8: o_1m_1 \neq o_3m_1, c_9: o_2m_1 \neq o_3m_1, \\~~~~~~~~~~~~c_{10}: o_1m_2 \neq o_2m_2, c_{11}: o_1m_2 \neq o_3m_2, \\~~~~~~~~~~~~c_{12}: o_2m_2 \neq o_3m_2, c_{13}: o_1m_3 \neq o_2m_3, \\~~~~~~~~~~~~c_{14}: o_1m_3 \neq o_3m_3, c_{15}: o_2m_3 \neq o_3m_3\}$
\item $S = \{s_1: o_1m_1=1, s_2: o_1m_2=2, s_3: o_1m_3=3, \\~~~~~~~~~~~s_4: o_2m_1=2, s_5: o_2m_2=3, s_6: o_2m_3=4, \\~~~~~~~~~~~s_7: o_3m_1=3, s_8: o_3m_2=4, s_9: o_3m_3=5\}$
\item $R_\rho = \{r_1': o_3m_3 < 5\}$
\end{itemize}

This reconfiguration task can be solved using \textsc{FlexDiag}. If we keep the ordering of the constraints as defined in $S$, \textsc{FlexDiag} (with $m=1$) returns the diagnosis $\Delta: \{s_1, s_2, s_3, s_7, s_8, s_9\}$  which can be used to determine the new solution $S' = \{s_1: o_1m_1=3, s_2: o_1m_2=4, s_3: o_1m_3=5, s_4: o_2m_1=2, s_5: o_2m_2=3, s_6: o_2m_3=4, s_7: o_3m_1=1, s_8: o_3m_2=2, s_9: o_3m_3=3\}$ (see Table \ref{schedulereconfiguration}). If we change the parametrization to $m=2$, \textsc{FlexDiag} returns the same diagnosis but in approximately half of the time (with 10 iterations, $16$ milliseconds were needed on an average for $m=2$ whereas $31$ milliseconds were needed for $m=1$). This is consistent with the estimates in Table \ref{numberconsistencychecks}.

Possible ordering criteria for constraints in such rescheduling scenarios can be, for example, customer value (changes related to orders of important customers should occur with a significantly lower probability) and the importance of individual orders. If some orders in a schedule should not be changed, this can be achieved by simply defining such requests as requirements ($R_\rho$), i.e., change requests as well as stability requests can be included as constraints $r_i'$ in $R_\rho$.

\begin{table}[ht]
\centering{}\begin{tabular}{|c|c|} 
\hline
$S$ & $S'$   \\
\hline
$s_1: o_1m_1=1$  & $s_1: o_1m_1=3$  \\
\hline
$s_4: o_2m_1 = 2$ & $s_4: o_2m_1 = 2$  \\
\hline
$s_7: o_3m_1 = 3$ & $s_7: o_3m_1 = 1$  \\
\hline
$s_2: o_1m_2 = 2$ & $s_2: o_1m_2 = 4$  \\
\hline
$s_5: o_2m_2 = 3$ & $s_5: o_2m_2 = 3$  \\
\hline
$s_8: o_3m_2 = 4$ & $s_8: o_3m_2 = 2$  \\
\hline
$s_3: o_1m_3 = 3$ & $s_3: o_1m_3 = 5$  \\
\hline
$s_6: o_2m_3 = 4$ & $s_6: o_2m_3 = 4$  \\
\hline
$s_9: o_3m_3 = 5$ & $s_9: o_3m_3 = 3$  \\
\hline
\end{tabular}   \caption{Reconfiguration determined for rescheduling task -- $S$ represents the original configuration and $S'$ represents a configuration resulting from a reconfiguration task.}  \label{schedulereconfiguration}
\end{table}

\section{Future Work}\label{FutureWork}

In our work, we focused on the evaluation of reconfiguration scenarios where the knowledge base itself is assumed to be consistent. In future work, we  will extend the \textsc{FlexDiag} algorithm to make it applicable in scenarios where knowledge bases are tested \cite{felfernig2004}. An example issue is to take into account situations where unintended configurations are accepted by the knowledge base. In this context, we will extend the work of \cite{felfernig2004} by not only taking into account negative test cases but also automatically generate relevant test cases, for example, on the basis of mutation testing approaches. We plan to extend our empirical evaluation to further industrial configuration knowledge bases. Furthermore, we want to analyze in which way we are able to further improve the output quality (e.g., in terms of minimality and accuracy) of \textsc{FlexDiag}, for example, by applying different constraint orderings depending on the observed interaction patterns (of users) and probability estimates for diagnosis membership derived thereof. The better potentially relevant constraints are predicted the better the diagnosis quality in terms of the mentioned metrics of minimality and accuracy. Note that, for example, counting the number of elements already identified as partial diagnosis elements in \textsc{FlexDiag} does not help to keep diagnosis determination within certain time limits, however, this mechanism could be used when determing more than one diagnosis to include diagnosis size as a relevance criterion. Also, in this context will analyze further alternatives to evaluate the quality of diagnoses which go beyond the metrics used in this article.

\section{Conclusions}\label{Conclusions}

Efficient reconfiguration functionalities are needed in various scenarios such as the reconfiguration of production schedules, the reconfiguration of the settings in mobile phone networks, and the reconfiguration of robot context information. We analyzed the \textsc{FlexDiag} algorithm with regard to potentials of improving existing direct diagnosis algorithms. When using  \textsc{FlexDiag}, there is a clear trade-off between performance of diagnosis calculation and diagnosis quality (measured, for example, in terms of minimality and accuracy).

\section*{Acknowledgements}
You can find the source code and material in \url{https://jagalindo.github.io/FlexDiag/} 

\bibliography{ecai2014}
\end{document}